\documentclass[journal,twoside,web]{ieeecolor}
\usepackage{cite}
\usepackage{amsmath,amssymb,amsfonts}
\usepackage{algorithmic}
\usepackage{graphicx}
\usepackage{textcomp}
\usepackage{adjustbox}
\let\labelindent\relax
\usepackage{enumitem}
\usepackage{float}
\usepackage{caption}
\usepackage{subcaption}
\usepackage{multirow}
\usepackage{mathtools}
\usepackage{commath}
\usepackage{booktabs}
\usepackage{bbm}
\usepackage[linesnumbered,ruled, vlined]{algorithm2e}

\def\logoname{LOGO-tmi-web}
\definecolor{subsectioncolor}{rgb}{0,0,0}
\def\BibTeX{{\rm B\kern-.05em{\sc i\kern-.025em b}\kern-.08em
    T\kern-.1667em\lower.7ex\hbox{E}\kern-.125emX}}
\begin{document}
\title{TransFair: Transferring Fairness from Ocular Disease Classification to Progression Prediction}

\author{Leila Gheisi, Chee-Hung Henry Chu, Raju Gottumukkala,  Yan Luo, Xingquan Zhu, Mengyu Wang, Min Shi
\thanks{Leila Gheisi, Chee-Hung Henry Chu, and Min Shi (corresponding author) are with the School of Computing and Informatics, University of Louisiana at Lafayette, LA, USA (e-mails: \{leila.gheisi1, henry.chu, min.shi\}@louisiana.edu).}
\thanks{Raju Gottumukkala is with the Informatics Research Institute, University of Louisiana at Lafayette, LA, USA (e-mail: raju.gottumukkala@louisiana.edu).}
\thanks{Xingquan Zhu is with the College of Engineering and Computer Science, Florida Atlantic University, FL, USA (e-mail: xzhu3@fau.edu).}
\thanks{Yan Luo, and Mengyu Wang are with the Harvard Ophthalmology AI Lab, Harvard Medical School, Boston, MA, USA (e-mail: \{yluo16, mengyu\_wang\}@meei.harvard.edu).}
}

\maketitle

% % define symbols used in the paper
% \newcommand{\modelname}{EyeLearn}
% \newcommand{\embeddim}{d}
% \newcommand{\ncluster}{C}
% \newcommand{\data}{\mathbf{D}}
% \newcommand{\datax}{\mathbf{X}}
% \newcommand{\datay}{\mathbf{Y}}
% \newcommand{\datasize}{|\mathbf{D}|}
% \newcommand{\spaceR}{\mathbf{R}}
% \newcommand{\embedspace}{\mathcal{H}}
% \newcommand{\embed}{\mathbf{h}}
% \newcommand{\weightw}{\mathbf{W}}
% \newcommand{\membank}{\mathbf{B}}
% \newcommand{\bankfeat}{\mathbf{V}}
% \newcommand{\bankclust}{\mathbf{C}}
% \newcommand{\width}{W}
% \newcommand{\lab}{c}
% \newcommand{\height}{H}
% \newcommand{\banksize}{M}
% \newcommand{\masks}{\mathbf{Z}}
% \newcommand{\maskm}{\mathbf{M}}
% \newcommand{\identity}{\mathbf{I}}
% \newcommand{\scaler}{r}
% \newcommand{\biasb}{b}
% \newcommand{\layerl}{l}
% \newcommand{\loss}{\mathcal{L}}
% \newcommand{\batchsize}{K}
% \newcommand{\dist}{\mathbf{u}}
% \newcommand{\batchembeds}{\mathbf{H}}
% \newcommand{\numnegs}{N}

\begin{abstract}
The use of artificial intelligence (AI) in automated disease classification significantly reduces healthcare costs and improves the accessibility of services. However, this transformation has given rise to concerns about the fairness of AI, which disproportionately affects certain groups, particularly patients from underprivileged populations. Recently, a number of methods and large-scale datasets have been proposed to address group performance disparities. Although these methods have shown effectiveness in disease classification tasks, they may fall short in ensuring fair prediction of disease progression, mainly because of limited longitudinal data with diverse demographics available for training a robust and equitable prediction model. In this paper, we introduce TransFair to enhance demographic fairness in progression prediction for ocular diseases. TransFair aims to transfer a fairness-enhanced disease classification model to the task of progression prediction with fairness preserved. Specifically, we train a fair EfficientNet, termed FairEN, equipped with a fairness-aware attention mechanism using extensive data for ocular disease classification. Subsequently, this fair classification model is adapted to a fair progression prediction model through knowledge distillation, which aims to minimize the latent feature distances between the classification and progression prediction models. We evaluate FairEN and TransFair for fairness-enhanced ocular disease classification and progression prediction using both two-dimensional (2D) and 3D retinal images. Extensive experiments and comparisons with models with and without considering fairness learning show that TransFair effectively enhances demographic equity in predicting ocular disease progression.
\end{abstract}


% \begin{keyword}
% AI fairness, disease progression, ocular disease, RNFLT maps, OCT B-scans
% \end{keyword}

% \end{frontmatter}

% holds immense promise for improving patient outcomes. However, ensuring fairness and equity in these AI models is paramount, as biases can perpetuate existing healthcare disparities. While progress has been made in addressing fairness in 2D medical imaging, the fairness of 3D models remains largely unexplored, primarily due to the scarcity of large-scale 3D fairness datasets. In this paper, we introduce TransFair, a novel approach that leverages knowledge distillation to transfer fairness from a teacher model trained on a large, diverse dataset for disease detection to a student model designed for progression prediction. We evaluate TransFair on the task of glaucoma progression prediction using 3D OCT scans from the Harvard-FairVision dataset, demonstrating its effectiveness in mitigating biases across race and gender while maintaining high predictive performance. Our work highlights the potential of transfer learning in promoting fairness in AI for healthcare, particularly in scenarios where data limitations hinder direct fairness interventions.

% \begin{IEEEkeywords}
% AI fairness, disease progression, ocular disease, fundus images, OCT B-scans
% \end{IEEEkeywords}

% \linenumbers

\section{Introduction}

% define symbols used in the paper
\newcommand{\modelname}{TransFair}
\newcommand{\modelfair}{FairEN}
\newcommand{\embeddim}{d}
\newcommand{\data}{\mathbf{D}}
\newcommand{\datax}{\mathbf{X}}
\newcommand{\attr}{\mathbf{A}}
\newcommand{\datasize}{N}
\newcommand{\datasizepro}{M}
\newcommand{\spaceR}{\mathbf{R}}
\newcommand{\embed}{\mathbf{h}}
\newcommand{\weight}{\mathbf{W}}
\newcommand{\score}{v}
\newcommand{\batchsize}{K}
\newcommand{\loss}{\mathcal{L}}

\begin{figure}
  \centering
    \includegraphics[width=0.47\textwidth]{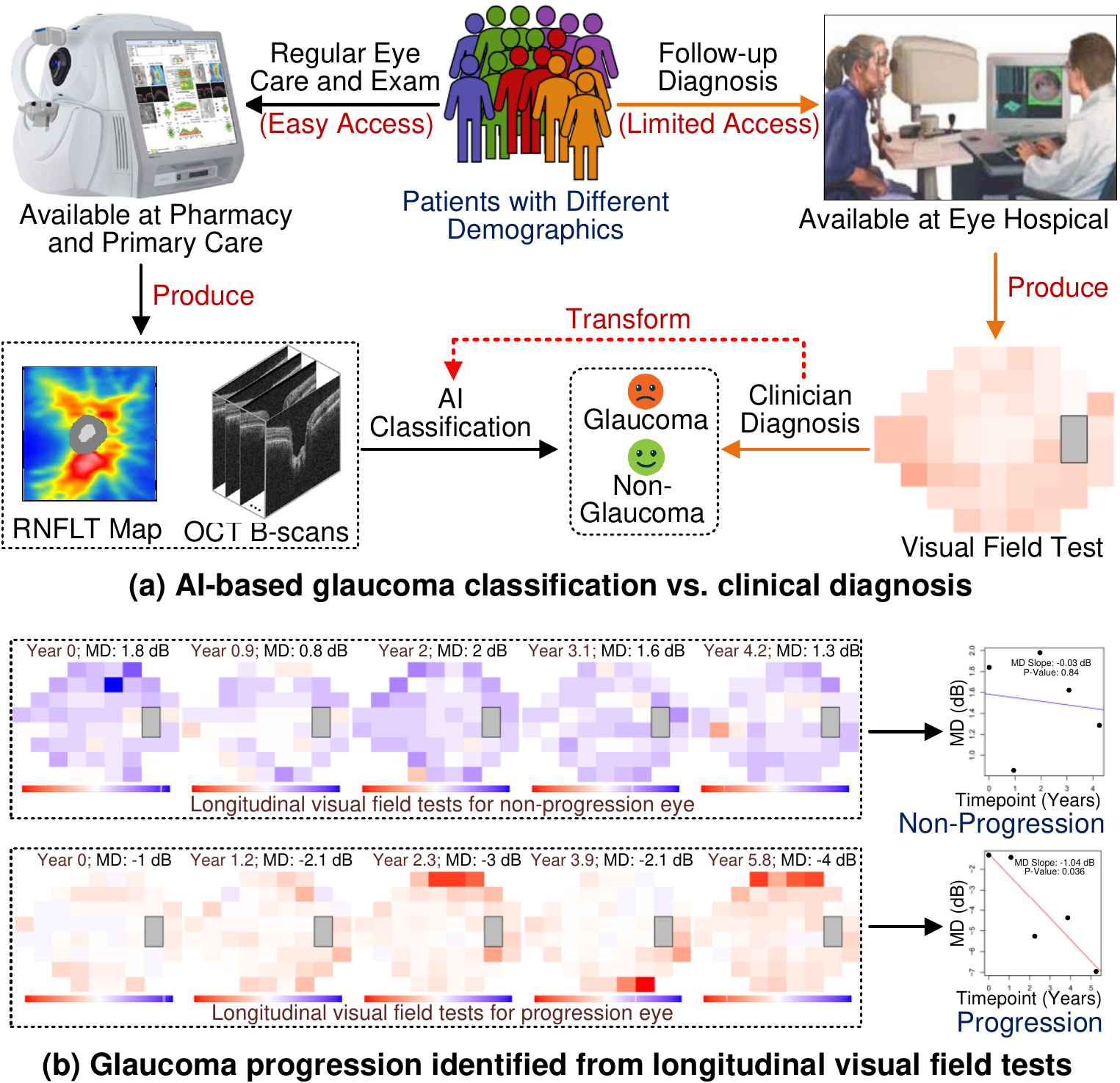}
  \caption{(a) AI-based glaucoma detection is more accessible and affordable compared to traditional clinician diagnoses. Patients can use the outcomes of AI detection as a basis for seeking further diagnosis from clinicians. (b) Longitudinal visual field tests are used to determine the progression of glaucoma. MD: mean deviation - the average of all the values in the visual field map.}
  \label{fig1}
\end{figure}

The rapid advancement of artificial intelligence (AI) technologies, particularly deep learning (DL), has profoundly reshaped the healthcare landscape, driving transformative changes throughout the industry \cite{acosta2022multimodal,zhang2024challenges}. 
This is demonstrated by the growing number of AI-enabled medical devices that have been approved by the United States Food and Drug Administration \cite{milam2023current,muehlematter2023fda}. Among these advancements, AI systems are extensively employed to aid in the detection of various diseases, achieving expert-level accuracies \cite{kumar2023artificial,tiu2022expert}. For example, deep learning methods are used for the automatic detection of glaucoma \cite{li2021applications,shi2024rnflt2vec}, as shown in Fig. \ref{fig1}a. Unlike traditional clinician diagnoses that are typically only accessible at hospitals, AI-based diagnostic systems can be implemented in local pharmacies and primary care settings for large-scale disease screening. This makes them more accessible and affordable for the general public, especially benefiting rural areas with limited clinical resources and economically disadvantaged populations. Although AI demonstrates remarkable abilities in disease classification, predicting disease progression is clinically more crucial due to the irreversible damage caused by certain conditions. For instance, ocular diseases like glaucoma can lead to permanent vision loss if preventive measures are not taken before its onset \cite{tham2014global}. Despite its critical importance, the task of progression prediction is currently less explored compared to classification, largely due to data scarcity \cite{li2023generating,liu2024imageflownet}. As shown in Fig. \ref{fig1}b, to accurately determine the progression or non-progression of glaucoma, longitudinal visual field data collected over extended periods are essential. However, collecting sufficient high-quality labeled data to train a reliable progression predictive model is prohibitively expensive and often unfeasible.

On the other hand, the application of DL has given rise to ethical debates and legal challenges in healthcare and medicine. A growing concern is the potential unfairness or bias of DL towards specific sub-populations \cite{chen2023algorithmic,luo2024harvard,tianfairseg}, particularly underprivileged groups defined by protected attributes like gender, race, ethnicity, socioeconomic status, among others. DL algorithms may identify spurious causal relationships in data that correlate with protected identity attributes, potentially leading to the use of patient identity information as a shortcut for predicting health outcomes \cite{pierson2021algorithmic}. For example, convolutional neural networks (CNN) disproportionately misdiagnose underserved groups, such as Hispanics and Medicaid patients, compared to White patients, and they inadvertently learn implicit racial information from radiology images \cite{glocker2023algorithmic}. In addition to above implicit mapping relationship learning, the unfairness of DL models can be caused by other confounding factors. First, the dataset utilized to train DL models can be imbalanced and skewed towards majority population groups. Consequently, DL models may perpetuate and amplify the inherent biases in the data, leading to unequal diagnostic outcomes for certain underserved populations. Second, the medical data annotations provided by specialists may contain noisy labels due to variations in clinical environments and the subjectivity of the specialists. Lastly, the models may exhibit bias towards simpler cases in diagnostic tasks, neglecting more challenging cases that are unevenly distributed across different sub-groups. Regardless of the implicit or explicit factors involved, it is crucial to ensure equitable diagnostic performance across different demographic groups before deploying an DL model in clinical practice to assist in the diagnosis.

In this work, we focus on improving the demographic fairness (e.g., across different racial and ethnic groups) of DL in predicting the progression of ocular diseases using retinal images. Research on fairness learning in disease progression prediction is scarce, largely due to the scarcity of medical data that includes comprehensive demographic attributes \cite{yuan2023algorithmic}. To the best of our knowledge, this is the first work to address fairness learning for predicting ocular disease progression. Mitigating the bias of DL in the medical settings is a complex and multifaceted challenge. A common approach is to manipulate the data used for training DL models to ensure they are representative of the entire population \cite{parraga2023fairness}. However, simply balancing the data distribution such as oversampling, synthetic data generation may be ineffective for certain medical applications \cite{shi2024equitable}. Another line of research concentrates on addressing algorithmic fairness, which involves introducing explicit fairness constraints to ensure equality in patient outcomes \cite{parraga2023fairness, chen2023algorithmic}. Given the robustness and generalizability of algorithmic-level fairness learning, we focus on improving the DL algorithmic fairness in this paper.

\begin{figure}
  \centering
    \includegraphics[width=0.47\textwidth]{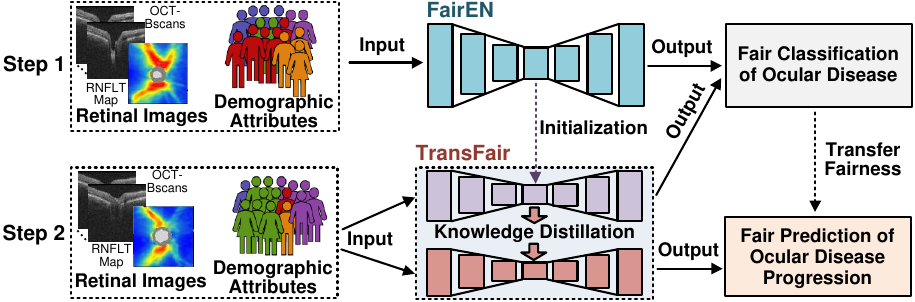}
  \caption{Illustration of the proposed \text{\modelname} framework trained in two steps. First, a \text{\modelfair} model (i.e. teacher) is trained to achieve fairness for classification. Then, the fair classification model guides the training of another \text{\modelfair} model (i.e. student) to achieve fairness for progression prediction. In \text{\modelname}, the  teacher and student models take the same inputs but perform classification and progression prediction tasks, respectively. The knowledge distillation aims to minimize the latent feature distances between the teacher and student models.}
  \label{fig2}
\end{figure}

To overcome the challenge of limited longitudinal retinal images with diverse demographics, we propose using a fair classification model trained on extensive data to enhance and guide the training of a fair progression model with limited data. Recently, several large-scale retinal image datasets for ocular disease classification, such as FairVision \cite{luo2023harvard} and FairDomain \cite{tian2024fairdomain}, have been developed within the community. These datasets facilitate the development of equitable DL models \cite{luo2023harvard,tian2024fairdomain} that enhance fairness in eye disease classification across various demographic groups defined by gender, race, and ethnicity. Motivated by these recent advancements, we aim to apply these insights and transfer demographic fairness from a classification model to a progression prediction model, as depicted in Fig. \ref{fig2}. To this end, we propose a novel fairness-aware feature learning model called \text{\modelfair}, which incorporates a demographic attribute-informed attention mechanism within the EfficientNet framework \cite{tan2019efficientnet} to tailor feature learning across diverse groups. Furthermore, we propose \text{\modelname}, a two-step fairness transfer approach from classification to progression prediction. First, we train a fairness-enhanced classification model (\text{\modelfair}) using sufficient retinal images. Subsequently, we utilize the pretrained fair classification model as a teacher to guide the training of the fair progression prediction model (\text{\modelfair} as a student) through knowledge distillation.

To conclude, our contributions are summarized as follows:
\begin{itemize}
    \item We propose to study a fair progression prediction problem for ocular disease, which is clinically vital for ensuring equitable progression prediction across various demographic groups.
    \item We propose \text{\modelfair}, which incorporates a fairness-aware attention mechanism into the EfficientNet to facilitate equitable classification and progression prediction of ocular diseases.
    \item We propose \text{\modelname}, which learns to transfer the demographic fairness property from a classification model to a progression prediction model with fairness preserved.
    \item We evaluate the proposed models for ocular disease classification and progression prediction using both 2D and 3D retinal images.
\end{itemize}

Section II surveys the related work. Section III presents preliminaries, including definition of the progression prediction problem and fairness learning basics used in our approach. The proposed model for fair progression prediction of ocular disease is introduced in section IV. Section V reports experimental results for accessing the proposed approach. Section VI discusses the proposed problem and solution. Finally, section VII concludes the paper.

\section{Related Work}
First, we review existing literature on ocular disease classification and progression prediction, and identify their limitations. Next, we explore recent advancements in fairness learning. Finally, we examine knowledge distillation technologies and their applications in the biomedical field.

\subsection{Ocular Disease Classification and Progression Prediction}
Ocular disease represents a significant global health issue, adversely affecting both daily activities and mental health. According to projections from the National Eye Institute \footnote{https://www.nei.nih.gov.}, by 2030, an estimated 4.3 million Americans will be diagnosed with glaucoma, 11.3 million with diabetic retinopathy, and 38.7 million with cataracts. These eye disorders can cause irreversible vision loss and potentially lead to blindness if not detected and treated promptly \cite{tham2014global,sarhan2020machine}. Over the past few years, deep learning-based approaches have been widely adopted for detecting various ocular diseases using retinal images such as color fundus photos, scanning laser ophthalmoscopy (SLO) fundus images, and optical coherence tomography (OCT) B-scans \cite{eladawi2018classification, badar2020application,sanghavi2024ocular}. For examples, Li et al. \cite{li2019large} proposed an CNN model with an attention mechanism that utilizes deep features highlighted by the visualized maps of pathological regions to automatically detect
glaucoma. Hemelings et al. \cite{hemelings2023generalizable} extended the CNN model for glaucoma screening from fundus images. Instead of a CNN that performs binary classification (glaucoma or not), they opted for a regression CNN that outputs a continuous risk score. This risk score for CNN training was expert estimated vertical cup-disc ratio, which increases alongside glaucoma severity. Yip et al. \cite{yip2020technical} analyzed various deep learning (DL) models for diabetic retinopathy detection, including VGGNet, ResNet, DenseNet, and Ensemble models. For diagnosing diabetic retinopathy, these four DL models demonstrated similar diagnostic effectiveness, with AUC values ranging from 0.936 to 0.944. Above methods mainly adopted a supervised training manner for eye disease classification. However, a major challenge is that accurate classification labels for large-scale retinal images are expensive and laborious to obtain. To mitigate this issue, Luo et al. proposed a novel generalization-reinforced semi-supervised learning model called pseudo supervisor to optimally utilize unlabeled data. The proposed pseudo supervisor optimizes the policy of predicting pseudo labels with unlabeled samples to improve empirical generalization.

Although numerous DL approaches have been developed that show promising performance in ocular disease classification, methods for predicting the progression of ocular diseases are less frequently explored \cite{thompson2020review}. Progression forecasting or prediction is a clinically more critical task for ocular patients compared to classification. It assists clinicians in determining the most appropriate treatment strategy—whether a patient should undergo aggressive treatment with invasive surgeries, which may have significant side effects, or opt for a more conservative approach using eye drops. Lee et al. \cite{lee2021predicting} developed a DL model called machine-to-machine to predict longitudinal changes of retinal nerve fiber layer thickness from fundus photographs. These changes were then used to predict future development of glaucomatous visual field defects using a joint longitudinal survival model. Arcadu et al. \cite{arcadu2019deep} developed a DL model that used color fundus to predict the future threat of significant DR worsening at a patient level over a span of two years after the baseline visit. They suggested that DL model would enable early identification of patients at highest risk of vision loss, allowing timely referral to retina specialists and potential initiation of treatment before irreversible vision loss occurs. 

\textit{Limitations of existing works:} 
While current studies have shown promising results in ocular disease classification and progression prediction, they face several practical challenges. First, current classification and progression prediction models often overlook the fairness aspect of DL, potentially leading to biased outcomes towards certain demographic groups \cite{parraga2023fairness}. Second, there is a scarcity of longitudinal retinal image data with diverse demographic attributes, which hinders the development of robust and equitable progression prediction models for ocular diseases. To address these challenges, we propose utilizing a fair classification model trained on extensive datasets to improve the training of a fair progression model using limited data.

\subsection{Fairness Learning}
A plethora of work has demonstrated that AI systems can exhibit bias in medical diagnostics, leading to uneven performance across various subgroups distinguished by protected attributes such as age, race, ethnicity, sex or gender, and socioeconomic status \cite{chen2023algorithmic,drukker2023toward}. The underlying factors of AI unfairness, which can be either explicit or implicit \cite{ricci2022addressing}, include factors such as data imbalance, noisy labels, and heterogeneity in anatomical and pathological features present in medical images across various demographic groups. 
Various techniques for mitigating bias to achieve demographic fairness have been introduced and can be categorized into pre-processing, in-processing, and post-processing methods \cite{chen2023algorithmic}. Pre-processing methods hypothesize that model bias is originated from the sample distribution imbalance, i.e., the majority of populations are from White and non-Hispanic groups compared to Asian and Black groups \cite{shi2024equitable}. Therefore, techniques such as data sampling and transfer learning, are commonly utilized to reduce model bias and enhance fairness \cite{qraitem2023bias}. In-processing methods are based on the hypothesis that deep learning models can inherit and perpetuate biases present in the data \cite{gichoya2022ai}. Therefore, effective de-biasing strategies must be incorporated during model training to mitigate the influence of demographic attributes. For example, adversarial training \cite{yang2023adversarial} is designed to prevent the model from learning identity-specific features from the images, thereby mitigating performance bias linked to identity. However, recent studies report that adversarial learning-based strategies may be ineffective for some medical applications, since demographic information can enhance performance for minority groups \cite{chen2023algorithmic,shi2024equitable}.
Luo et al. \cite{luo2024harvard} proposed a fair identity normalization technique to normalize the latent image features of different groups with learnable means and standard deviations. As a result of this normalization, different groups will learn more distinguishing features benefiting improved model performance and equity. Rather than training a fair model from scratch, post-processing methods modify the output of a trained model (such as probability scores or decision thresholds) to satisfy group-fairness metrics \cite{chen2023algorithmic}. Xian et al. \cite{xian2024optimal} proposed a post-processing algorithm for fair classification under parity group fairness criteria, applicable to binary and multi-class problems under both attribute-aware and the more general attribute-blind setting. Gennaro et al. \cite{di2024post} framed the post-processing task as a new supervised learning problem taking as input the previously trained (biased) model. They introduced a new approach, Ratio-Based Model Debiasing, which predicts a multiplicative factor to rescale the biased model’s predictions such that they better satisfy fairness guarantees. IN this work, we focus on the in-processing fairness learning method given that it is robust and generalizable in different applicable scenarios.

\textit{Differences from existing works:} Existing fairness learning studies primarily focus on natural images or tabular data with simplified attributes, such as gender, for fairness learning. These differ from the more complex tasks present in healthcare and medical settings. First, medical AI systems must accommodate a range of demographic groups, including various racial and ethnic categories. Second, the pathological features in medical images often display subtle differences across these groups, posing a challenge in training both discriminative and fair deep learning models. Recently, several works have been proposed to address the fairness issue for medical classification \cite{tian2024fairdomain,luo2024harvard, ktena2024generative}. In contrast, our focus in this paper is on fairness learning for disease progression prediction tasks, which are more challenging due to issues with data scarcity.

\subsection{Knowledge Distillation in Biomedical Applications}
Knowledge distillation is a technique for model compression where a well-trained \cite{meng2021knowledge}, larger model (the teacher model) is used to guide a smaller model (the student model) to achieve outcomes that closely approximate those of the larger model. According to the number of teacher, Knowledge distillation can be roughly classified into three categories: single teacher knowledge distillation, multi-teacher knowledge distillation, and no teacher knowledge distillation (commonly known as self distillation). While according to the category of distilled knowledge, knowledge distillation can be divided into three categories: logit-based, feature-based, and relationship-based. In this work, we focus on single-teacher and feature-based knowledge distillation to transfer a fair classification model to a progression prediction model with fairness preserved. Knowledge distillation has been widely used in the medical field \cite{meng2021knowledge}. Wu et al. \cite{wu2024adaptive} proposed an adaptive knowledge distillation approach for unsupervised MRI reconstruction. The teacher models were firstly trained by filling the re-undersampled images and compared with the undersampled images in a self-supervised manner.The teacher models are then distilled to train another cascade model that can leverage the entire undersampled k-space during its training and testing. Xing et al. proposed a novel class-guided contrastive distillation module to pull closer positive image pairs from the same class in the teacher and student models, while pushing apart negative image pairs from different classes. With this regularization, the feature distribution of the student model shows higher intra-class similarity and inter-class variance \cite{xing2021categorical}. 
Unlike the above methods, in our work, the teacher model is trained for classification and the student model for progression prediction, with both models achieving demographic fairness.

% \subsection{Glaucoma and OCT B-scans}
% Glaucoma is the second leading cause of blindness worldwide and a major public health concern in the United States. Today, people face a wide range of diseases, making it essential to either find cures or detect these conditions early to enable prevention or treatment. Glaucoma is a common eye disorder and a leading cause of blindness. This disease gradually damages the eye's blood vessels, ultimately resulting in vision loss.\cite{chauhan2016data}Glaucoma is the second leading cause of blindness after cataracts, making its early detection crucial to prevent vision loss.\cite{chauhan2012proposed} Early detection plays a critical role in managing glaucoma. Optical Coherence Tomography (OCT) enables the detection of subtle structural changes caused by glaucoma, offering a wealth of parameters that provide comprehensive insights into the eye’s condition.\cite{wu2021comparison}

\begin{figure*}[t]
  \centering
    \includegraphics[width=0.98\textwidth]{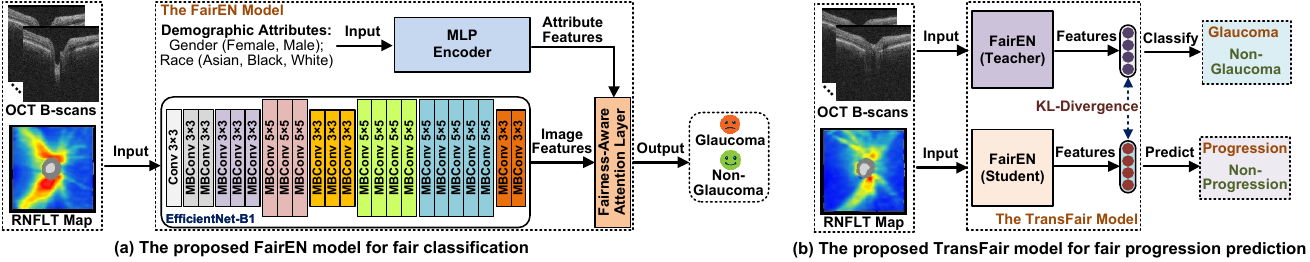}
  \caption{The proposed approach for fairness-enhanced progression prediction of ocular disease using retinal images. (a) Train a fair classification model based on the \text{\modelfair}. (b) Transfer fairness from classification to progression prediction based on the \text{\modelname}.}
  \label{fig3}
\end{figure*}

\section{Problem Definition and Preliminaries}
This section explains key concepts, problem formulation, evaluation tasks, and foundational knowledge related to the proposed method.

\noindent\textbf{RNFLT Map and OCT B-scan:} This work utilizes two types of retinal images for the classification and progression prediction tasks, including 2D retinal nerve fiber layer thickness (RNFLT) map and 3D optical coherence tomography (OCT) B-scans. RNFLT is a measurement that quantifies the distance between the internal limiting membrane and the outer aspect of the retinal nerve fiber layer. Each RNFLT map captures the thickness values within the peripapillary area of $6 \times 6$ mm$^2$. RNFLT maps are instrumental in diagnosing conditions such as glaucoma and other retinal disorders. Swept-source OCT is a non-invasive imaging technique which generates cross-sectional images of the retinal with high resolution. OCT B-scans are useful for diagnosing ocualr diseases such as glaucoma, diabetic retinopathy and macular degeneration.

\noindent\textbf{Ocular Disease Classification:} Let $\data_{cls} \in \spaceR^{\datasize}$ represent a dataset of $\datasize$ retinal images for classification, which may be either 2D or 3D. Each image belongs to a specific population group (e.g., Asians, Blacks, Whites) defined by a demographic attribute $\attr_{cls}$ (e.g., race). The DL model is trained to learn the latent features of each retinal image, $\datax_{cls} \in \data_{cls}$, expressed as $\embed_{cls} = f_{cls}(\datax_{cls})$. This training occurs in a supervised manner by classifying types of ocular diseases, such as glaucoma and non-glaucoma. In this work, the DL model indicates the proposed \text{\modelfair}.

\noindent\textbf{Ocular Disease Progression Prediction:} Let $\data_{pred} \in \spaceR^{\datasizepro}$ represent a dataset of $\datasizepro$ retinal images. Each image $\datax_{pred} \in \data_{pred}$ belongs to a subgroup defined by the demographic attribute $\attr_{pred}$. Similar to the classification task, a DL model (i.e., \text{\modelname}) is trained to learn the latent features $\embed_{pred}$ of each retinal image. This training occurs in a supervised manner by predicting the progression outcome (i.e., progression vs. non-progression) of ocular diseases. The binary progression label is identified from longitudinal retinal images of multiple diagnoses over an extended period of time. 

\noindent\textbf{Demographic Fairness of DL Models:} The goal of the DL models is to fairly learn the latent features (i.e., $\embed_{cls}$ and $\embed_{pred}$) of retinal images, thereby ensuring equitable classification and progression prediction of ocular diseases across various demographic groups. These groups include gender categories such as Females and Males, or racial categories such as Asians, Blacks, and Whites. In this study, demographic fairness is achieved when these different groups exhibit equal or statistically indistinguishable performances in tasks related to ocular disease classification and progression prediction.

\section{Methodology}
This section explains the proposed approach in Fig. \ref{fig3} for fairness-enhanced progression prediction of the ocular disease. It involves the following three model training components in two consecutive phases: 
\begin{itemize}
    \item \textbf{Classification of Disease based on \text{\modelfair} (Phase 1):} A fairness-enhanced classification model is trained based on the \text{\modelfair} model to achieve enhanced demographic fairness.  
    \item \textbf{Progression Prediction of Disease based on \text{\modelname} (Phase 2):} The fair classification model guides the training of a fair progression prediction model by training two sub-networks in \text{\modelname}: \textbf{(a)} The classification model from phase 1 serves as a teacher and is further optimized for the disease classification task; \textbf{(b)} Under the guidance of this fair classification model through knowledge distillation, a student model is trained to carry out progression prediction with fairness preserved.
\end{itemize}

It is important to note that in phase 2, both the teacher and student models are \text{\modelfair} and they process the same retinal images, albeit for different tasks: classification and progression prediction, respectively. The details of these models and their optimization processes are introduced in the subsequent sections.

\subsection{Classification of Disease based on \text{\modelfair}}
We aim to train a fair classification model on a substantial set of retinal images (e.g., RNFLT maps and OCT B-scans), that represent diverse demographics. This model will then guide the training of a fair progression prediction model, which will operate on a more limited dataset. To achieve this, we introduce a fairness-aware version of EfficientNet \cite{tan2019efficientnet}, referred to as \text{\modelfair}, as shown in Fig. \ref{fig3}a. 

\text{\modelfair} uses EfficientNet-B1 (written as EfficientNet thereafter) as the backbone to learn features of each retinal image $ \datax_{cls} \in \spaceR^{\datasize}$, represented as:
\begin{equation}
    \embed_{cls} = \textit{EfficientNet-B1}(\datax_{cls})
\end{equation}
The EfficientNet architecture comprises seven blocks differentiated by color in Fig. \ref{fig3}a, each consisting of multiple MBConv layers—a mobile inverted bottleneck convolution design that integrates squeeze-and-excitation optimization. These blocks vary in the depth and size of their filters as indicated by the kernel sizes (3 $\times$ 3 or 5 $\times$ 5) and the number of layers in each block. Starting with the initial standard convolution layer, the network progresses through these MBConv blocks, which increase in complexity and depth. Each block represents a stage in feature extraction where similar operations are applied to the input features. This hierarchical structure allows EfficientNet to efficiently manage computational resources while maximizing learning and representational capacity, making it highly effective for tasks requiring detailed image analysis and classification.

However, EfficientNet may exhibit biases towards certain demographic groups (e.g., Asians and Blacks), resulting in notable performance discrepancies among different populations. To address this, we introduce a fairness-aware attention mechanism that adjusts feature learning based on demographic attributes. As shown in Fig. \ref{fig3}a, we utilize a multilayer perceptron (MLP) encoder to process the demographic attributes $\attr_{cls}$ associated with the input image $\datax_{cls}$:
\begin{equation}
    \embed^{cls}_{attr} = \textit{MLP}(\attr_{cls})
\end{equation}
Subsequently, the image and demographic attribute features are used to compute the fairness-aware attention weight $\score$ by:
\begin{equation}
    \embed_q = \embed^{cls}_{attr}\weight^1, \ \embed_k = \embed_{cls}\weight^2
\end{equation}
\begin{equation}
    v = softmax(\frac{\embed_q\embed_k}{\sqrt{\embeddim}})
\end{equation}
where $\embeddim$ is the dimension of latent features, $\weight^1$ and $\weight^2$ are learnable weight parameters. Next, the output of the fairness-aware attention layer is computed by:
\begin{equation}
    \embed_{cls} = \embed_{cls}\weight^3 \cdot v
\end{equation}
where $\weight^3$ is the weight parameter. Finally, $\embed_{cls}$ is used for classification via a liner mapping layer which generates a probability value.

\subsection{Progression Prediction of Disease based on \text{\modelname}}
We introduce \text{\modelname}, designed to transition a fairness-enhanced classification model into a progression prediction model while maintaining its fairness properties. As shown in Fig. \ref{fig3}b, \text{\modelname} includes a teacher \text{\modelfair} and a student \text{\modelfair}, which performs different tasks co-trained through the process of knowledge distillation.

\subsubsection{Optimize the Teacher Model for Classification} The teacher model fine-tunes the fairness-enhanced classification model that was pretrained in the previous section. The teacher \text{\modelfair} is optimized for both performance and equity, utilizing either RNFLT maps or OCT B-scans for the supervised classification of the ocular disease. Since both the pretraining and finetuning of the classification model are conducted on the same type of retinal images, there is no domain shift involved. Consequently, it is expected that the teacher model will maintain and even improve its demographic fairness. Similar to the pretraining phase, the teacher model trains to minimize the binary classification loss. 

\subsubsection{Train the Student Model for Progression Prediction} We train another \text{\modelfair} model as a student for disease progression prediction, as depicted in Fig. \ref{fig3}b. The student model takes the retinal image $\datax_{pred} \in \spaceR^{\datasizepro}$ and demographic attribute $\attr_{pred}$ as inputs and learns respective image and attribute features as follows:
\begin{equation}
     \embed_{pred} = \textit{EfficientNet-B1}(\datax_{pred})
\end{equation}

\begin{equation}
     \embed_{attr}^{pred} = \textit{MLP}(\attr_{pred})
\end{equation}
Similar to teacher model's feature learning process, the learned image features after the fairness-aware attention layer are represented by:
\begin{equation}
    \embed_{pred} = \embed_{pred}\weight^6\cdot softmax(\frac{\embed^{pred}_{attr}\weight^4\embed_{pred}\weight^5}{\sqrt{\embeddim}})
\end{equation}
where $\weight^4$, $\weight^5$ and $\weight^6$ are learnable weight parameters. To enhance the feature learning and fairness in the student model, image and attribute feature similarities between teacher and student models are minimized based on the Kullback-Leibler (KL) divergence \cite{joyce2011kullback}:
\begin{equation}
    D^{img}_{KL} (\embed_{cls}\parallel \embed_{pred}) = \frac{1}{\batchsize} \sum \embed_{cls} \log \left(\frac{\embed_{cls}}{\embed_{pred}} \right)
\end{equation}
\begin{equation}
    D^{attr}_{KL} (\embed^{cls}_{attr}\parallel \embed^{pred}_{attr}) = \frac{1}{\batchsize} \sum \embed^{cls}_{attr} \log \left(\frac{\embed^{cls}_{attr}}{\embed^{pred}_{attr}} \right)
\end{equation}
where $\batchsize$ is the batch size during the training. Taken together, the knowledge distillation minimizes:
\begin{equation}
    D_{KL} = \alpha \cdot D^{img}_{KL} + \beta \cdot D^{attr}_{KL}
\end{equation}
where $\alpha$ and $\beta$ are hyper-parameters used to control the image and attribute feature similarities between teacher and student models. With this optimization, we anticipate that both the feature learning capacity and fairness attributes of the classification model will be transferred to the progression prediction model.

\SetInd{0.48em}{0.77em}
\begin{algorithm}[t]
\begin{small}
\DontPrintSemicolon
\SetAlgoLined
\SetKwInOut{Input}{Input}\SetKwInOut{Output}{Output}
\Input{The dataset $\datax_{cls}$ with $\datasize$ retinal images, respective demographic attributes, and classification labels} 
\Output{ocular disease classification}
Initialize dimension $\embeddim$, weight parameters $\weight^1$, $\weight^2$ and $\weight^3$, training epochs $I_{cls}$, and batch size $\batchsize$.\\
\BlankLine
\For {$i \in [1,I_{cls}]$}{
    \For{$j \in [1,\datasize // \batchsize]$}{
    $\embed_{cls}\leftarrow$ learn features of image $\datax^j_{cls}$ based on Eq. 1;\\
    $\embed_{attr}^{cls}\leftarrow$ learn features of demographic attribute $\attr^j_{cls}$ based on Eq. 2;\\
    $v\leftarrow$ calculate fairness-aware attention weight based on Eq. 4; \\
    $\embed_{cls}\leftarrow$ obtain fairness-aware image features based on Eq. 5.
    }
    $\loss_{cls}\leftarrow$ calculate the classification loss for $\batchsize$ images in batch $j$ based on Eq. 12. \\
    Minimize $\loss_{cls}$ based on the Adam optimizer. 
}
\caption{Training of \text{\modelfair} for classification}
\label{tab1}
\end{small}
\end{algorithm}

\SetInd{0.48em}{0.77em}
\begin{algorithm}[t]
\begin{small}
\DontPrintSemicolon
\SetAlgoLined
\SetKwInOut{Input}{Input}\SetKwInOut{Output}{Output}
\Input{The dataset $\datax_{pred}$ with $\datasizepro$ retinal images, demographic attributes, classification and progression labels} 
\Output{ocular disease progression outcome} 
Initialize dimension $\embeddim$, weight parameters $\weight^4$, $\weight^5$ and $\weight^6$, training epochs $I_{pred}$, batch size $\batchsize$, and the teacher model.\\
\BlankLine
\For {$i \in [1,I_{pred}]$}{
    \For{$j \in [1,\datasizepro // \batchsize]$}{
    $\embed_{cls}\leftarrow$ obtain fairness-aware image features from the teacher model based on Eq. 5; \\
    $\embed_{pred}\leftarrow$ obtain fairness-aware image features from the student model based on Eq. 8; \\
    }
    $D^{img}_{KL}\leftarrow$ obtain image feature similarity based on Eq. 9;\\
    $D^{attr}_{KL}\leftarrow$ obtain demographic attribute feature similarity based on Eq. 10;\\
    $D_{KL}\leftarrow$ obtain image and attribute feature similarities based on Eq. 11;\\
    $\loss_{cls}\leftarrow$ calculate the classification loss for $\batchsize$ images in batch $j$ based on Eq. 12. \\
    $\loss_{pred}\leftarrow$ calculate the progression prediction loss for $\batchsize$ images in batch $j$ based on Eq. 13. \\
    Minimize $\loss_{cls}$ and $\loss_{pred}$ based on the AdamW optimizer. 
}
\label{tab2}
\caption{Training of \text{\modelname} for progression prediction}
\end{small}
\end{algorithm}

\subsection{Optimizations for \text{\modelfair} and \text{\modelname}} 
\text{\modelfair} and \text{\modelname} are trained in supervised manners. \text{\modelfair} aims to minimize the classification loss, which is a binary cross-entropy loss over $\batchsize$ retinal images in a mini-batch as:
\begin{equation}
    \loss_{cls} = -\frac{1}{\batchsize}\sum_j [y_j\cdot \text{log}\sigma(\hat{y}_j) + (1 - y_j) \cdot \text{log}(1 - \sigma(\hat{y}_j))]
\end{equation}
where $y_i$ and $\hat{y}_i$ are the ground-truth and predicted classification labels for current image $\datax^j_{cls}$. The detailed training process of \text{\modelfair} is summarized in Algorithm \ref{tab1}. \text{\modelname} optimize both teacher and student models. The teacher model has the same optimization loss in Eq. 12, while the student model aims to minimize a combined loss:
% \begin{equation}
%     \loss_{pred} = & -\frac{1}{\batchsize}\sum_j [z_j\cdot \text{log}\sigma(\hat{z}_j) + (1 - z_j) \cdot \text{log}(1 - \sigma(\hat{z}_j))] \\
%     & + D_{KL}
% \end{equation}
\begin{align}
    \loss_{pred} = & -\frac{1}{\batchsize} \sum_j \big[z_j \cdot \text{log}\sigma(\hat{z}_j) \nonumber \\
    & + (1 - z_j) \cdot \text{log}(1 - \sigma(\hat{z}_j)) \big] \nonumber \\
    & + D_{KL}
\end{align}
where $z_i$ and $\hat{z}_i$ are the ground-truth and predicted progression labels for current image $\datax^j_{cls}$. $D_{KL}$ is calculated through Eq. 11. The detailed training process is summarized in Algorithm \ref{tab2}.

\section{EVALUATION METHODS AND RESULTS}
This section presents experiments and comparative analyses to assess the performance and fairness of the proposed \text{\modelfair} and \text{\modelname} models using both 2D and 3D retinal images. 

\subsection{Datasets}
In this study, we focus on glaucoma, the second leading cause of blindness worldwide, although our approach is extensible to other diseases. Our evaluation of model fairness is centered on two demographic attributes: gender, categorized as Females and Males, and race, including Asians, Blacks, and Whites. We utilize three datasets to assess our models: Harvard-GF \cite{luo2024harvard}, FairVision \cite{luo2023harvard}, and Harvard-GDP \cite{luo2023harvard}. The details of these datasets are as follows:
\begin{itemize}[leftmargin=9pt]
    \item  \textbf{Harvard-GF:} It consists of 3,300 RNFLT maps from 3,300 patients, with each 2D RNFLT map having dimensions of 225 $\times$ 225. Self-reported demographic information includes a gender distribution of 54.9\% female and 45.1\% male, and a racial breakdown of 33.3\% White, 33.3\% Black, and 33.4\% Asian. Within the dataset, 47.0\% of the patients are classified as non-glaucoma and 53.0\% as glaucoma.
    \item \textbf{FairVision:} It includes 10,000 OCT data samples from 10,000 patients. Each 3D OCT data sample contains 200 B-scans, where each B-scan image has a dimension of 200 $\times$ 200. The self-reported patient demographic information is as follows: 57.0\% of the patients are female and 43.0\% are male; racially, 8.5\% are Asian, 14.9\% are Black, and 76.6\% are White. Glaucoma classification is based on a thorough clinical assessment. Accordingly, 51.3\% of the patients are categorized as non-glaucoma and 48.7\% as glaucoma.
    \item \textbf{Harvard-GDP:} It includes 500 data samples from 500 glaucoma patients. Each sample includes both RNFLT map and OCT B-scans, accompanied by labels for glaucoma classification and progression prediction. Demographic information is as follows: Gender distribution is 54.0\% female and 46.0\% male; racial composition includes 9.4\% Asian, 15.6\% Black, and 75.0\% White. The patients are divided into non-glaucoma and glaucoma categories, making up 55.2\% and 44.8\% of the dataset, respectively. Two criteria are employed to determine glaucoma progression based on visual field maps \cite{vesti2003comparison}, with each map represented by a vector of 52 total deviation (TD) values ranging from -38 dB to 26 dB (see Fig. \ref{fig1}b for reference). The criteria are: (1) MD Fast Progression: eyes with an MD slope $\leq$ -1 dB. (2) TD Progression: eyes with at least three locations exhibiting a TD slope $\leq$ -1 dB;  According to MD fast progression, 91.2\% are categorized as non-progression and 8.8\% as progression. For TD progression, the figures are 90.6\% for non-progression and 9.4\% for progression.
\end{itemize}

We perform experiments using 2D RNFLT maps and 3D OCT B-scans independently. The Harvard-GF and FairVision datasets are utilized for pretraining the fairness-enhanced classification model (as detailed in Section III-A), whereas the Harvard-GDP dataset is employed to train \text{\modelname} for fairness-aware progression prediction (as described in Section III-B).

\subsection{Comparative Methods}
We choose to compare with the following methods for progression prediction with or without considering fairness learning:

\noindent\textbf{\textit{Without fairness learning:}}
\begin{itemize}[leftmargin=9pt]
    \item \textbf{VGG} \cite{simonyan2014very}: The VGG model is a deep convolutional neural network that uses small convolution filters to enhance depth and performance in image recognition tasks.
    \item \textbf{DenseNet} \cite{huang2017densely}: DenseNet connects each layer to every other layer in a feed-forward fashion, promoting feature reuse and reducing the number of parameters.
    \item \textbf{ResNet} \cite{he2016deep}: ResNet is a deep convolutional neural network that uses skip connections in its residual blocks to train deeper networks effectively, addressing the vanishing gradient problem.
    \item \textbf{ViT} \cite{dosovitskiy2020image}: Vision Transformer (ViT) applies the transformer architecture to image processing by treating patches of an image as sequences for powerful, scalable image recognition.
    \item \textbf{EfficientNet} \cite{tan2019efficientnet}: It is a scalable deep convolutional neural network that optimizes accuracy and efficiency by scaling up layers, width, and resolution based on a compound coefficient.
\end{itemize}

\noindent\textbf{\textit{With fairness learning:}}
\begin{itemize}[leftmargin=9pt]
    \item \textbf{EfficientNet\_Adv} \cite{beutel2017data}: It integrates an adversarial training to prevent learning demographic attribute features in EfficientNet.
    % \item \textbf{EfficientNet + FIS} \cite{luo2023harvard}: It integrates a fair identity scaling component that enables EfficientNet to adjust its feature learning according to demographic attributes.
    \item \textbf{\modelfair:} The proposed \text{\modelfair} which incorporates a fairness-aware attention mechanism into the EfficientNet.
    \item \textbf{\modelname:} The proposed \text{\modelname} for fairness-enhanced progression prediction of the ocular disease.
\end{itemize}

\subsection{Experimental Settings}
\noindent\textbf{Dataset Splits:} To train \text{\modelfair} for glaucoma detection, we follow the data separation as the original paper \cite{luo2023harvard}, where 70\% of the samples are allocated for training, 10\% and 20\% are used for model evaluation and testing, respectively. To Train \text{\modelname} for the progression prediction task, 70\% and 30\% are used for training and testing, respectively. We focus on gender and race for demographic attributes.

\noindent\textbf{Training Scheme:} The CNN models including VGG, DenseNet, ResNet and EfficientNet are trained with a learning rate of 1e-4 for 10 epochs with a batch size of 6. For ViT, we follow the literature \cite{he2022masked} settings and train for 50 epochs with a layer decay of 0.55, weight decay of 0.01, dropout rate of 0.1, batch size of 64, and base learning rate of 5e-4. AdamW \cite{loshchilov2017decoupled} is used as the optimization method for all models. The default values for $\alpha$ and $\beta$ are set as 1.0 and 0.05 in \text{\modelname}.

\begin{figure}
  \centering
    \includegraphics[width=0.47\textwidth]{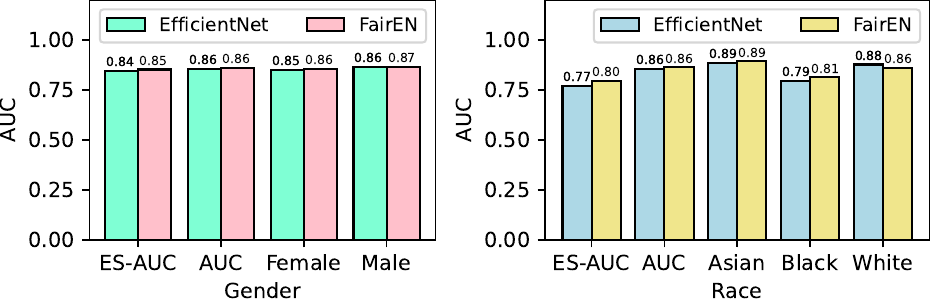}
    \vspace{-3mm}
  \caption{The glaucoma detection performance using RNFLT maps.}
  \label{fig4}
\end{figure}

\begin{table*}[t]
\centering
\caption{The MD Fast Progression Prediction using RNFLT Maps.}
\small
\renewcommand{\arraystretch}{1}
\setlength{\tabcolsep}{3pt}
\begin{tabular}{@{}lcccccccccc@{}}
\toprule
\textbf{} & \multicolumn{4}{c}{\textbf{Gender}} & \multicolumn{5}{c}{\textbf{Race}} \\ 
\cmidrule(lr){2-5} \cmidrule(lr){6-10}
\textbf{Model} & \textbf{ES-AUC} & \textbf{Overall AUC} & \textbf{Female AUC} & \textbf{Male AUC} & \textbf{ES-AUC} & \textbf{Overall AUC} & \textbf{Asian AUC} & \textbf{Black AUC} & \textbf{White AUC} \\
\midrule
VGG              & 0.5452 & 0.6095 & 0.5687 & 0.6866 & 0.4311 & 0.6095 & 0.9286 & 0.5714 & 0.5528 \\
DenseNet         & 0.5447 & 0.6488 & 0.5851 & 0.7761 & 0.4590 & 0.6488 & 0.8571 & 0.4524 & 0.6400 \\
ResNet           & 0.5867 & 0.6498 & 0.6119 & 0.7194 & 0.5139 & 0.6498 & 0.6429 & 0.4286 & 0.6860 \\
ViT              & 0.6814 & 0.6960 & 0.7060 & 0.7075 & 0.5622 & 0.6960 & 0.6429 & 0.8571 & 0.6722 \\
EfficientNet     & 0.4946 & 0.6806 & 0.5791 & 0.9552 & 0.6158 & 0.6806 & 0.7143 & 0.6190 & 0.6905 \\
\midrule
EfficientNet\_Adv & 0.4808 & 0.6473 & 0.5254 & 0.8716 & 0.4559 & 0.6448 & 1.0000 & 0.6429 & 0.5877 \\
\textbf{\modelfair} & 0.6405 & 0.7189 & 0.7552 & 0.6328 & 0.5022 & 0.7045 & 0.8929 & 0.8571 & 0.6428 \\
\textbf{\modelname} & 0.7131 & 0.7716 & 0.7418 & 0.8239 & 0.5714 & 0.7975 & 0.9286 & 0.9524 & 0.6878 \\
\bottomrule
\end{tabular}
\label{tab1}
\end{table*}

\begin{table*}[t]
\centering
\caption{The TD Pointwise Progression Prediction using RNFLT Maps.}
\small
\renewcommand{\arraystretch}{1}
\setlength{\tabcolsep}{3pt}
\begin{tabular}{@{}lcccccccccc@{}}
\toprule
\textbf{} & \multicolumn{4}{c}{\textbf{Gender}} & \multicolumn{5}{c}{\textbf{Race}} \\ 
\cmidrule(lr){2-5} \cmidrule(lr){6-10}
\textbf{Model} & \textbf{ES-AUC} & \textbf{Overall AUC} & \textbf{Female AUC} & \textbf{Male AUC} & \textbf{ES-AUC} & \textbf{Overall AUC} & \textbf{Asian AUC} & \textbf{Black AUC} & \textbf{White AUC} \\
\midrule
VGG              & 0.6672 & 0.7270 & 0.6796 & 0.7692 & 0.5172 & 0.7270 & 0.8000 & 1.0000 & 0.6675 \\
DenseNet         & 0.6572 & 0.7376 & 0.6755 & 0.7978 & 0.5373 & 0.7376 & 0.8000 & 1.0000 & 0.6896 \\
ResNet           & 0.7161 & 0.7360 & 0.7531 & 0.7253 & 0.5535 & 0.7360 & 0.8667 & 0.9091 & 0.7100 \\
ViT              & 0.5653 & 0.5942 & 0.6224 & 0.5714 & 0.3791 & 0.5942 & 0.7333 & 0.9545 & 0.5264 \\
EfficientNet     & 0.6260 & 0.7725 & 0.6694 & 0.9033 & 0.6107 & 0.7725 & 0.7333 & 0.9545 & 0.7287 \\
\midrule
EfficientNet\_Adv & 0.6146 & 0.7788 & 0.6449 & 0.9121 & 0.5621 & 0.7693 & 0.8667 & 1.0000 & 0.7287 \\
\textbf{\modelfair} & 0.6228 & 0.7693 & 0.6571 & 0.8923 & 0.6204 & 0.7852 & 0.8000 & 1.0000 & 0.7491 \\
\textbf{\modelname} & 0.7119 & 0.7794 & 0.7469 & 0.8418 & 0.6136 & 0.7984 & 0.8667 & 1.0000 & 0.7670 \\
\bottomrule
\end{tabular}
\label{tab2}
\end{table*}

\noindent\textbf{Evaluation Metrics:} We evaluate \text{\modelfair} for binary glaucoma detection and evaluate \text{\modelname} for binary progression prediction based on MD fast progression and TD progression, respectively. We use the Area Under the Receiver Operating Characteristic Curve (AUC) to access the glaucoma detection and progression prediction performance. To access model fairness, we use equity-scaled AUC following the existing work \cite{luo2023harvard}, which is computed by:
\begin{equation*}
    \begin{split}
    \text{ES-AUC} = \frac{\text{AUC}}{1+\sum_{a}^{\mathcal{A}}| \text{AUC} - \text{AUC}_{a} |}
\end{split}
\label{eqn:es_metric}
\end{equation*}
where $\mathcal{A}$ is the set of subgroups (e.g., {Asian, Black, White}) for a specific demographic attribute (e.g., race). AUC is the overall model performance on all patients, while AUC$_a$ indicates AUC for subgroup $a$ for the demographic attribute $\mathcal{A}$.

\subsection{Results}
This section presents the results of \text{\modelfair} and \text{\modelname}, comparing their performance and fairness with existing strong methods.

\subsubsection{Performance of \text{\modelfair} for Classification} In this work, we aim to use a fairness-enhanced detection model to guide the training of a fair progression prediction model for ocular diseases. Using 2D RNFLT maps, \text{\modelfair} has generally improved both glaucoma detection performance and fairness, as shown in Fig. \ref{fig4}. For the gender attribute, the ES-AUC improved from 0.84 to 0.85 and the overall AUC remains the same. This demonstrates that \text{\modelfair} improved the model fairness without harming the overall performance, which are attributed by the AUC improvement of 0.01 for both Female and Male groups. For the racial attribute, the overall AUC and ES-AUC improved by 0.01 and 0.03, respectively, which again means the model fairness has been enhanced after incorporating the fairness-aware attention mechanism into the EfficientNet.

\begin{figure}
  \centering
    \includegraphics[width=0.47\textwidth]{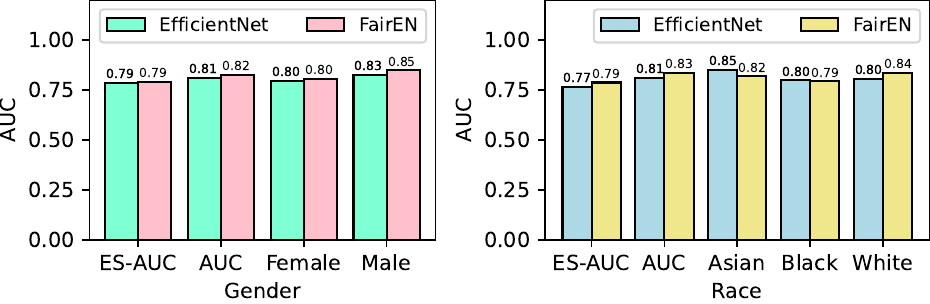}
    \vspace{-3mm}
  \caption{The glaucoma detection performance using OCT B-scans.}
  \label{fig5}
\end{figure}

Using 3D OCT B-scans for glaucoma detection, both AUC and ES-AUC improved by 0.01 for the gender attribute, as shown in Fig. \ref{fig5}. The improvement is more prominent for the racial attribute, with AUC and ES-AUC both improved by 0.02. Taken together, above observations demonstrate that \text{\modelfair} is able to enhance the model fairness using both 2D and 3D retinal images without compromising the overall detection performance. The fairness-enhanced glaucoma detection model will be used the boost the model fairness and performance for predicting glaucoma progression.

\begin{table*}[t]
\centering
\caption{The MD Fast Progression Prediction using OCT B-scans.}
\small
\renewcommand{\arraystretch}{1}
\setlength{\tabcolsep}{3pt}
\begin{tabular}{@{}lcccccccccc@{}}
\toprule
\textbf{} & \multicolumn{4}{c}{\textbf{Gender}} & \multicolumn{5}{c}{\textbf{Race}} \\ 
\cmidrule(lr){2-5} \cmidrule(lr){6-10}
\textbf{Methods} & \textbf{ES-AUC} & \textbf{Overall AUC} & \textbf{Female AUC} & \textbf{Male AUC} & \textbf{ES-AUC} & \textbf{Overall AUC} & \textbf{Asian AUC} & \textbf{Black AUC} & \textbf{White AUC} \\
\midrule
VGG              & 0.6295 & 0.6493 & 0.6433 & 0.6239 & 0.4866 & 0.6493 & 0.3571 & 0.6667 & 0.6740 \\
DenseNet         & 0.6453 & 0.6771 & 0.6970 & 0.6478 & 0.5383 & 0.6771 & 0.7143 & 0.8571 & 0.6364 \\
ResNet           & 0.6686 & 0.6886 & 0.6896 & 0.6597 & 0.4905 & 0.6886 & 1.0000 & 0.6429 & 0.6419 \\
ViT              & 0.6055 & 0.6498 & 0.6224 & 0.6955 & 0.5352 & 0.6498 & 0.6071 & 0.8095 & 0.6382 \\
EfficientNet     & 0.6302 & 0.6622 & 0.6448 & 0.6955 & 0.5553 & 0.6622 & 0.5714 & 0.5952 & 0.6970 \\
\midrule
EfficientNet\_Adv & 0.5771 & 0.6831 & 0.7448 & 0.5612 & 0.5658 & 0.7642 & 0.8929 & 0.5714 & 0.7934 \\
\textbf{\modelfair} & 0.5404 & 0.6856 & 0.6269 & 0.8955 & 0.5787 & 0.7627 & 0.6786 & 0.5714 & 0.8053 \\
\textbf{\modelname} & 0.6931 & 0.7383 & 0.7537 & 0.7881 & 0.6768 & 0.7781 & 0.8571 & 0.7143 & 0.7713 \\
\bottomrule
\end{tabular}
\label{tab3}
\end{table*}

\begin{table*}[t]
\centering
\caption{The TD Pointwise Progression Prediction using OCT B-scans.}
\small
\renewcommand{\arraystretch}{1}
\setlength{\tabcolsep}{3pt}
\begin{tabular}{@{}lcccccccccc@{}}
\toprule
\textbf{} & \multicolumn{4}{c}{\textbf{Gender}} & \multicolumn{5}{c}{\textbf{Race}} \\ 
\cmidrule(lr){2-5} \cmidrule(lr){6-10}
\textbf{Methods} & \textbf{ES-AUC} & \textbf{Overall AUC} & \textbf{Female AUC} & \textbf{Male AUC} & \textbf{ES-AUC} & \textbf{Overall AUC} & \textbf{Asian AUC} & \textbf{Black AUC} & \textbf{White AUC} \\
\midrule
VGG              & 0.6650  & 0.6889 & 0.7041 & 0.6681 & 0.5473 & 0.6889 & 0.6667 & 0.9091 & 0.6726 \\
DenseNet         & 0.6539 & 0.6831 & 0.7061 & 0.6615 & 0.4627 & 0.6831 & 0.9333 & 0.8636 & 0.6378 \\
ResNet           & 0.5210  & 0.5693 & 0.5204 & 0.6132 & 0.4029 & 0.5693 & 0.8000 & 0.7273 & 0.5451 \\
ViT              & 0.5202 & 0.5884 & 0.6673 & 0.5363 & 0.3847 & 0.5884 & 0.7333 & 0.9091 & 0.5247 \\
EfficientNet     & 0.7036 & 0.7185 & 0.7245 & 0.7033 & 0.5340 & 0.7185 & 0.6667 & 0.4545 & 0.7483 \\
\midrule
EfficientNet\_Adv & 0.6200  & 0.7079 & 0.7857 & 0.6440 & 0.6003 & 0.7257 & 0.6667 & 0.5909 & 0.7406 \\
\textbf{\modelfair} & 0.6605 & 0.7206 & 0.6694 & 0.7604 & 0.5069 & 0.7344 & 0.6000 & 1.0000 & 0.7832 \\
\textbf{\modelname} & 0.7194 & 0.7434 & 0.7673 & 0.7341 & 0.6452 & 0.7460 & 0.6667 & 0.8182 & 0.7509 \\
\bottomrule
\end{tabular}
\label{tab4}
\end{table*}

\subsubsection{Performance of \text{\modelname} for Progression Prediction} We aim to enhance fairness in progression prediction by utilizing the fairness-enhanced ocular disease detection model built on the proposed \text{\modelname} framework. Table \ref{tab1} shows the MD fast progression prediction using RNFLT maps. We can have the following five major observations. First, among all CNN methods, EfficientNet performns the best for both gender and racial attributes in terms of the overall AUC. For example, EfficientNet improved the AUC by 0.0711 over VGG, 0.0328 over DenseNet, and 0.0308 over ResNet. In addition, EfficientNet has the best model fairness on race, although it demonstrates the least fair CNN model. Second, ViT generally performs better than EfficientNet in both model performance and fairness in progression prediction using RNFLT maps, especially on gender where the overall AUC and ES-AUC outperform by 0.0154 and 0.1868, respectively. Third, integrating the adversarial training into EfficientNet does not enhance the model performance and fairness. Adversarial training aims to prevent the preservation of demographic attribute features for progression prediction. However, the comparison between EfficientNet and EfficientNet\_Adv illustrates that the sensitive attribute information might be beneficial for enhancing both overall model performance and equity. Forth, both the overall AUC and ES-AUC of \text{\modelfair} are significantly higher than those of EfficientNet and Efficient\_Adv, which demonstrates the effectiveness of \text{\modelfair} for fairness-enhanced progression prediction. For example, for the gender attribute, the AUC and ES-AUC of \text{\modelfair} improved by 0.0383 and 0.1459 over EfficientNet, and 0.0716 and 0.1597 over EfficientNet\_Adv, respectively. Last, among all fairness learning methods, \text{\modelname} performs the best in model performance and fairness for both gender and racial attributes as observed from Table \ref{tab1}. The significant performance and fairness improvements are attributed to the supervision of the glaucoma detection model as a teacher. In other words, the well-trained teacher model using extensive retinal images for glaucoma detection is able to boost the performance of the student model for progression prediction using limited dataset. Additionally, the fairness-enhanced glaucoma detection model is helpful to boost the model fairness in progression prediction.

The experimental results on RNFLT maps for TD pointwise progression are shown in Table \ref{tab2}. We can observe that EfficientNet achieves the best overall AUC among all CNN methods. In addition, it has the best ES-AUC for the racial attribute, although with smallest ES-AUC for the gender attribute. However, EfficientNet can be considered as the best CNN model because the oversall AUC is the first priority to consider in this case, i.e., it is undesirable to improve model fairness by sacrificing the overall model performance. The AUC and ES-AUC of \text{\modelfair} both improved by 0.01 for the racial attribute but do not perform significantly better than EfficientNet. Overall, \text{\modelfair} outperforms EfficientNet\_Adv in overall performance and fairness for both demographic attributes. This again demonstrate that \text{\modelfair} is effective to boost both model performance and fairness for progression prediction. Furthermore, the \text{\modelname} model leveraging the fairness-enhanced glaucoma detection model as a teacher to supervise the progression prediction model can generally improve the overall performance and model fairness. For example, compared with \text{\modelfair}, the AUC and -ES-AUC of \text{\modelname} improved by 0.01 and 0.0891 for the gender attribute. 

Similar observations can be obtained on OCT B-scans for MD progression, as depicted in Table \ref{tab3}. \text{\modelfair} generally performs better than EfficientNet and EfficientNet\_Adv for both gender and racial attributes. For instance, the overall AUC of \text{\modelfair} significantly improved by 0.02 on gender and 0.10 on race. This demonstrate the proposed fair attention mechanism is useful to enhance the performance of progression prediction. The fair attention also enhanced the ES-AUC by 0.02 on the racial attribute. On both demographic attributes, \text{\modelname} performs the best regarding model performance and fairness among all comparative methods. This verifies the argument that a fairness-enhanced ocular disease classification model can be leveraged to enhance model performance and fairness in the progression prediction task. This conclusion can be further evidenced by the comparative results on TD pointswise progress from Table \ref{tab4}. From Table \ref{tab4}, we can observe that \text{\modelfair} performs better than EfficientNet and EfficientNet\_Adv in overall AUC for both attributes, i.e., improved by 0.0159 and 0.0087 over EfficientNet and EfficientNet\_Adv on race, and 0.0021 and 0.0127 over EfficientNet and EfficientNet\_Adv on gender. \text{\modelname} still remains the best model in AUC and ES-AUC for both attributes among all comparative methods, as shown in Table \ref{tab4}.

\subsection{Parameter Sensitivity}
\begin{figure}
  \centering
    \includegraphics[width=0.47\textwidth]{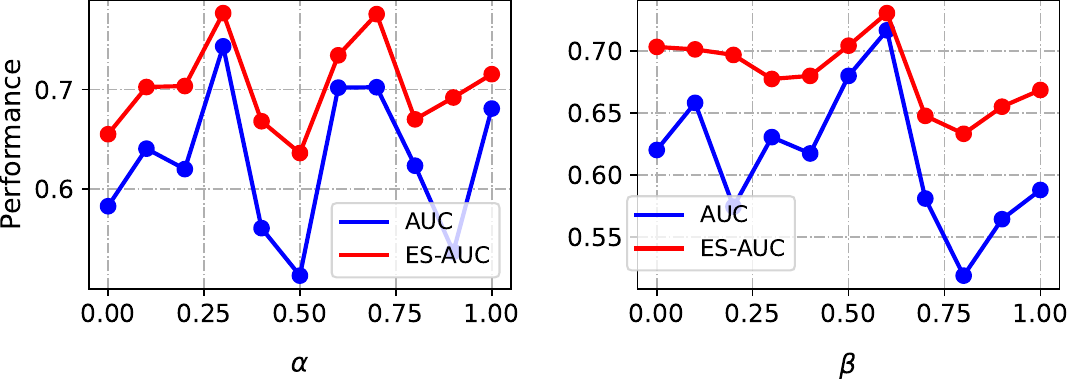}
    \vspace{-3mm}
  \caption{The impacts of $\alpha$ and $\beta$ for the gender attribute on RNFLT maps.}
  \label{fig6}
\end{figure}

\begin{figure}
  \centering
    \includegraphics[width=0.47\textwidth]{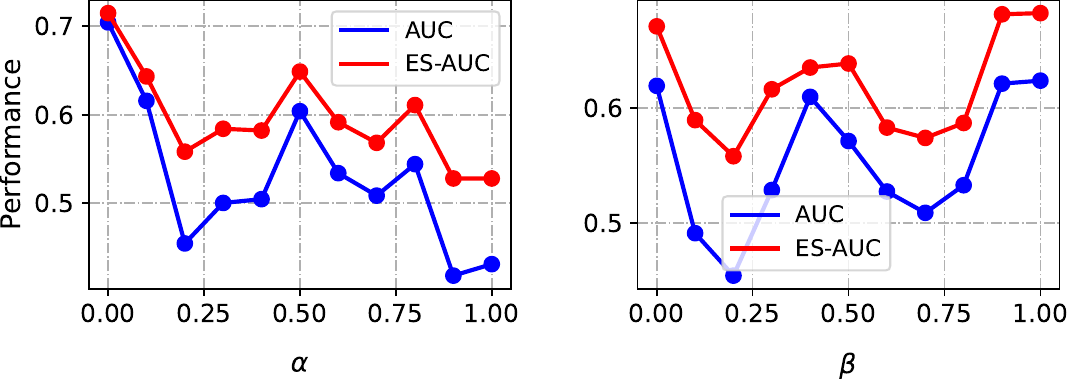}
    \vspace{-3mm}
  \caption{The impacts of $\alpha$ and $\beta$ for the gender attribute on OCT B-scans.}
  \label{fig7}
\end{figure}

We evaluate the sensitivities of the two key parameters, $\alpha$ and $\beta$, which regulate the image and attribute feature similarities between the teacher and student models in \text{\modelname}. 
A larger value of $\alpha$ or $\beta$ indicates that the knowledge distillation between the teacher and student models places greater emphasis on image similarity or attribute feature similarity, respectively. When analyzing the impact of varying $\alpha$ or $\beta$, the other parameter is fixed at its default value of 0.2. Fig. \ref{fig6} demonstrates the impacts of the two parameters for the gender attribute on RNFLT maps. Fig. \ref{fig7} demonstrates the impacts of the two parameters for the gender attribute on OCT B-scans. We can observe that the AUC and ES-AUC vary significantly from the changes of $\alpha$ and $\beta$. Therefore, it is important to choose a proper value setting for $\alpha$ and $\beta$ in order to achieve best model performance and fairness.

\section{Discussion}
Progression prediction of diseases is crucial for treatment planning, such as to determine if aggressive or conservative treatment is needed, in the clinical settings. However, training a responsive especially an demographically equitable deep learning model for progression prediction is mainly hindered by the scarcity of longitudinal data to observe the progression status of diseases. We propose to mitigate this challenge by leveraging a fairness-enhanced disease classification model trained with extensive data to boost the model performance and fairness in progression prediction with limited data. This is based on the hypothesis the learning ability and fairness acquired from one model can be transferred to another model through the knowledge distillation. Extensive experiments using both 2D and 3D medical images verified that it is feasible to achieve fairness-enhanced progression prediction with the guidance of a fairness-enhanced disease classification model. 

Our work have several limitations. First, the experiments mainly focus on retinal images and eye disease, while the proposed models have not been evaluated for other human diseases due to a lack of relevant longitudinal image data with demographic attributes. We plan to extend our evaluation to more diseases once relevant datasets are publicly available in the community. Second, we mainly focus on the EfficientNet to design the fairness learning model in this work because its efficiency compared with other CNN models. However, the proposed fairness attention mechanism is easily extensible to combine with other models, which can be a future direction. Last, we design the fairness attention mechanism as add-on layers after the EfficientNet feature learning layers. Despite its simplicity, we have not tested the model when the fairness learning layers are added after each learning block in the EfficientNet architecture, which could be a more effective fairness learning mechanism.

\section{Conclusion}
% In this work, we propose to transfer the demographic fairness property from a classification model to a progression prediction model for ocular diseases. We propose \text{\modelfair} with fairness-aware feature learning for fairness-enhanced ocular disease classification using extensive retinal images of glaucoma patients. Furthermore, we develop \text{\modelname} with fair knowledge distillation which treat the fairness-enhanced classification model as a teacher to supervise the training of a progression prediction model. To evaluate the propose models, we design extensive experiments and comparisons with existing methods with or without considering fairness learning using both 2D and 3D retinal images from three different real-word datasets. We use AUC and ES-AUC to access the model performance and fairness, respectively. The experimental results demonstrate \text{\modelfair} is able to boost both model performance and fairness for gender and racial attributes in the ocular disease classification task. The experiments also verify that \text{\modelname} is effective to improve the demographic fairness of the model in the progression prediction task. The proposed models are generalizable to other diseases for equitable progression prediction.
In this work, we propose a novel approach to transfer the demographic fairness property from a classification model to a progression prediction model for ocular diseases. Specifically, we introduce \text{\modelfair}, a fairness-aware feature learning framework for enhancing fairness in ocular disease classification using extensive retinal images of glaucoma patients. Additionally, we develop \text{\modelname}, which employs fair knowledge distillation by leveraging the fairness-enhanced classification model as a teacher to guide the training of a progression prediction model. To evaluate the proposed models, we conduct extensive experiments and comparisons with existing methods, both with and without fairness considerations, using 2D and 3D retinal images from three diverse real-world datasets. Model performance and fairness are assessed using AUC and ES-AUC metrics, respectively. Experimental results demonstrate that \text{\modelfair} improves both performance and fairness for gender and racial attributes in ocular disease classification tasks. Moreover, \text{\modelname} effectively enhances demographic fairness in the progression prediction task. The proposed models are generalizable and can be applied to other diseases for equitable progression prediction.

% \clearpage

\bibliographystyle{IEEEtran}
\bibliography{references}

\begin{thebibliography}{10}
\providecommand{\url}[1]{#1}
\csname url@samestyle\endcsname
\providecommand{\newblock}{\relax}
\providecommand{\bibinfo}[2]{#2}
\providecommand{\BIBentrySTDinterwordspacing}{\spaceskip=0pt\relax}
\providecommand{\BIBentryALTinterwordstretchfactor}{4}
\providecommand{\BIBentryALTinterwordspacing}{\spaceskip=\fontdimen2\font plus
\BIBentryALTinterwordstretchfactor\fontdimen3\font minus \fontdimen4\font\relax}
\providecommand{\BIBforeignlanguage}[2]{{%
\expandafter\ifx\csname l@#1\endcsname\relax
\typeout{** WARNING: IEEEtran.bst: No hyphenation pattern has been}%
\typeout{** loaded for the language `#1'. Using the pattern for}%
\typeout{** the default language instead.}%
\else
\language=\csname l@#1\endcsname
\fi
#2}}
\providecommand{\BIBdecl}{\relax}
\BIBdecl

\bibitem{acosta2022multimodal}
J.~N. Acosta, G.~J. Falcone, P.~Rajpurkar, and E.~J. Topol, ``Multimodal biomedical ai,'' \emph{Nature Medicine}, vol.~28, no.~9, pp. 1773--1784, 2022.

\bibitem{zhang2024challenges}
S.~Zhang and D.~Metaxas, ``On the challenges and perspectives of foundation models for medical image analysis,'' \emph{Medical image analysis}, vol.~91, p. 102996, 2024.

\bibitem{milam2023current}
M.~Milam and C.~Koo, ``The current status and future of fda-approved artificial intelligence tools in chest radiology in the united states,'' \emph{Clinical Radiology}, vol.~78, no.~2, pp. 115--122, 2023.

\bibitem{muehlematter2023fda}
U.~J. Muehlematter, C.~Bluethgen, and K.~N. Vokinger, ``Fda-cleared artificial intelligence and machine learning-based medical devices and their 510 (k) predicate networks,'' \emph{The Lancet Digital Health}, vol.~5, no.~9, pp. e618--e626, 2023.

\bibitem{kumar2023artificial}
Y.~Kumar, A.~Koul, R.~Singla, and M.~F. Ijaz, ``Artificial intelligence in disease diagnosis: a systematic literature review, synthesizing framework and future research agenda,'' \emph{Journal of ambient intelligence and humanized computing}, vol.~14, no.~7, pp. 8459--8486, 2023.

\bibitem{tiu2022expert}
E.~Tiu, E.~Talius, P.~Patel, C.~P. Langlotz, A.~Y. Ng, and P.~Rajpurkar, ``Expert-level detection of pathologies from unannotated chest x-ray images via self-supervised learning,'' \emph{Nature Biomedical Engineering}, vol.~6, no.~12, pp. 1399--1406, 2022.

\bibitem{li2021applications}
T.~Li, W.~Bo, C.~Hu, H.~Kang, H.~Liu, K.~Wang, and H.~Fu, ``Applications of deep learning in fundus images: A review,'' \emph{Medical Image Analysis}, vol.~69, p. 101971, 2021.

\bibitem{shi2024rnflt2vec}
M.~Shi, Y.~Tian, Y.~Luo, T.~Elze, and M.~Wang, ``Rnflt2vec: Artifact-corrected representation learning for retinal nerve fiber layer thickness maps,'' \emph{Medical Image Analysis}, vol.~94, p. 103110, 2024.

\bibitem{tham2014global}
Y.-C. Tham, X.~Li, T.~Y. Wong, H.~A. Quigley, T.~Aung, and C.-Y. Cheng, ``Global prevalence of glaucoma and projections of glaucoma burden through 2040: a systematic review and meta-analysis,'' \emph{Ophthalmology}, vol. 121, no.~11, pp. 2081--2090, 2014.

\bibitem{li2023generating}
J.~Li, B.~J. Cairns, J.~Li, and T.~Zhu, ``Generating synthetic mixed-type longitudinal electronic health records for artificial intelligent applications,'' \emph{NPJ Digital Medicine}, vol.~6, no.~1, p.~98, 2023.

\bibitem{liu2024imageflownet}
C.~Liu, K.~Xu, L.~L. Shen, G.~Huguet, Z.~Wang, A.~Tong, D.~Bzdok, J.~Stewart, J.~C. Wang, L.~V. Del~Priore \emph{et~al.}, ``Imageflownet: forecasting multiscale image-level trajectories of disease progression with irregularlysampled longitudinal medical images,'' \emph{arXiv preprint arXiv}, vol. 2406, p.~13, 2024.

\bibitem{chen2023algorithmic}
R.~J. Chen, J.~J. Wang, D.~F. Williamson, T.~Y. Chen, J.~Lipkova, M.~Y. Lu, S.~Sahai, and F.~Mahmood, ``Algorithmic fairness in artificial intelligence for medicine and healthcare,'' \emph{Nature biomedical engineering}, vol.~7, no.~6, pp. 719--742, 2023.

\bibitem{luo2024harvard}
Y.~Luo, Y.~Tian, M.~Shi, L.~R. Pasquale, L.~Q. Shen, N.~Zebardast, T.~Elze, and M.~Wang, ``Harvard glaucoma fairness: a retinal nerve disease dataset for fairness learning and fair identity normalization,'' \emph{IEEE Transactions on Medical Imaging}, 2024.

\bibitem{tianfairseg}
Y.~Tian, M.~Shi, Y.~Luo, A.~Kouhana, T.~Elze, and M.~Wang, ``Fairseg: A large-scale medical image segmentation dataset for fairness learning using segment anything model with fair error-bound scaling,'' in \emph{The Twelfth International Conference on Learning Representations}, 2024.

\bibitem{pierson2021algorithmic}
E.~Pierson, D.~M. Cutler, J.~Leskovec, S.~Mullainathan, and Z.~Obermeyer, ``An algorithmic approach to reducing unexplained pain disparities in underserved populations,'' \emph{Nature Medicine}, vol.~27, no.~1, pp. 136--140, 2021.

\bibitem{glocker2023algorithmic}
B.~Glocker, C.~Jones, M.~Bernhardt, and S.~Winzeck, ``Algorithmic encoding of protected characteristics in chest x-ray disease detection models,'' \emph{EBioMedicine}, vol.~89, 2023.

\bibitem{yuan2023algorithmic}
C.~Yuan, K.~A. Linn, and R.~A. Hubbard, ``Algorithmic fairness of machine learning models for alzheimer disease progression,'' \emph{JAMA Network Open}, vol.~6, no.~11, pp. e2\,342\,203--e2\,342\,203, 2023.

\bibitem{parraga2023fairness}
O.~Parraga, M.~D. More, C.~M. Oliveira, N.~S. Gavenski, L.~S. Kupssinsk{\"u}, A.~Medronha, L.~V. Moura, G.~S. Sim{\~o}es, and R.~C. Barros, ``Fairness in deep learning: A survey on vision and language research,'' \emph{ACM Computing Surveys}, 2023.

\bibitem{shi2024equitable}
M.~Shi, M.~M. Afzal, H.~Huang, C.~Wen, Y.~Luo, M.~O. Khan, Y.~Tian, L.~Kim, T.~Elze, Y.~Fang \emph{et~al.}, ``Equitable deep learning for diabetic retinopathy detection using multi-dimensional retinal imaging with fair adaptive scaling: a retrospective study,'' \emph{medRxiv}, pp. 2024--04, 2024.

\bibitem{luo2023harvard}
Y.~Luo, Y.~Tian, M.~Shi, T.~Elze, and M.~Wang, ``Harvard eye fairness: A large-scale 3d imaging dataset for equitable eye diseases screening and fair identity scaling,'' \emph{arXiv e-prints}, pp. arXiv--2310, 2023.

\bibitem{tian2024fairdomain}
Y.~Tian, C.~Wen, M.~Shi, M.~M. Afzal, H.~Huang, M.~O. Khan, Y.~Luo, Y.~Fang, and M.~Wang, ``Fairdomain: Achieving fairness in cross-domain medical image segmentation and classification,'' \emph{arXiv preprint arXiv:2407.08813}, 2024.

\bibitem{tan2019efficientnet}
M.~Tan and Q.~Le, ``Efficientnet: Rethinking model scaling for convolutional neural networks,'' in \emph{International conference on machine learning}.\hskip 1em plus 0.5em minus 0.4em\relax PMLR, 2019, pp. 6105--6114.

\bibitem{sarhan2020machine}
M.~H. Sarhan, M.~A. Nasseri, D.~Zapp, M.~Maier, C.~P. Lohmann, N.~Navab, and A.~Eslami, ``Machine learning techniques for ophthalmic data processing: a review,'' \emph{IEEE Journal of Biomedical and Health Informatics}, vol.~24, no.~12, pp. 3338--3350, 2020.

\bibitem{eladawi2018classification}
N.~Eladawi, M.~Elmogy, M.~Ghazal, O.~Helmy, A.~Aboelfetouh, A.~Riad, S.~Schaal, and A.~El-Baz, ``Classification of retinal diseases based on oct images,'' \emph{Front Biosci}, vol.~23, no.~2, pp. 247--264, 2018.

\bibitem{badar2020application}
M.~Badar, M.~Haris, and A.~Fatima, ``Application of deep learning for retinal image analysis: A review,'' \emph{Computer Science Review}, vol.~35, p. 100203, 2020.

\bibitem{sanghavi2024ocular}
J.~Sanghavi and M.~Kurhekar, ``Ocular disease detection systems based on fundus images: a survey,'' \emph{Multimedia Tools and Applications}, vol.~83, no.~7, pp. 21\,471--21\,496, 2024.

\bibitem{li2019large}
L.~Li, M.~Xu, H.~Liu, Y.~Li, X.~Wang, L.~Jiang, Z.~Wang, X.~Fan, and N.~Wang, ``A large-scale database and a cnn model for attention-based glaucoma detection,'' \emph{IEEE transactions on medical imaging}, vol.~39, no.~2, pp. 413--424, 2019.

\bibitem{hemelings2023generalizable}
R.~Hemelings, B.~Elen, A.~K. Schuster, M.~B. Blaschko, J.~Barbosa-Breda, P.~Hujanen, A.~Junglas, S.~Nickels, A.~White, N.~Pfeiffer \emph{et~al.}, ``A generalizable deep learning regression model for automated glaucoma screening from fundus images,'' \emph{NPJ digital medicine}, vol.~6, no.~1, p. 112, 2023.

\bibitem{yip2020technical}
M.~Y. Yip, G.~Lim, Z.~W. Lim, Q.~D. Nguyen, C.~C. Chong, M.~Yu, V.~Bellemo, Y.~Xie, X.~Q. Lee, H.~Hamzah \emph{et~al.}, ``Technical and imaging factors influencing performance of deep learning systems for diabetic retinopathy,'' \emph{NPJ digital medicine}, vol.~3, no.~1, p.~40, 2020.

\bibitem{thompson2020review}
A.~C. Thompson, A.~A. Jammal, and F.~A. Medeiros, ``A review of deep learning for screening, diagnosis, and detection of glaucoma progression,'' \emph{Translational vision science \& technology}, vol.~9, no.~2, pp. 42--42, 2020.

\bibitem{lee2021predicting}
T.~Lee, A.~A. Jammal, E.~B. Mariottoni, and F.~A. Medeiros, ``Predicting glaucoma development with longitudinal deep learning predictions from fundus photographs,'' \emph{American journal of ophthalmology}, vol. 225, pp. 86--94, 2021.

\bibitem{arcadu2019deep}
F.~Arcadu, F.~Benmansour, A.~Maunz, J.~Willis, Z.~Haskova, and M.~Prunotto, ``Deep learning algorithm predicts diabetic retinopathy progression in individual patients,'' \emph{NPJ digital medicine}, vol.~2, no.~1, p.~92, 2019.

\bibitem{drukker2023toward}
K.~Drukker, W.~Chen, J.~Gichoya, N.~Gruszauskas, J.~Kalpathy-Cramer, S.~Koyejo, K.~Myers, R.~C. S{\'a}, B.~Sahiner, H.~Whitney \emph{et~al.}, ``Toward fairness in artificial intelligence for medical image analysis: identification and mitigation of potential biases in the roadmap from data collection to model deployment,'' \emph{Journal of Medical Imaging}, vol.~10, no.~6, pp. 061\,104--061\,104, 2023.

\bibitem{ricci2022addressing}
M.~A. Ricci~Lara, R.~Echeveste, and E.~Ferrante, ``Addressing fairness in artificial intelligence for medical imaging,'' \emph{nature communications}, vol.~13, no.~1, p. 4581, 2022.

\bibitem{qraitem2023bias}
M.~Qraitem, K.~Saenko, and B.~A. Plummer, ``Bias mimicking: A simple sampling approach for bias mitigation,'' in \emph{Proceedings of the IEEE/CVF Conference on Computer Vision and Pattern Recognition}, 2023, pp. 20\,311--20\,320.

\bibitem{gichoya2022ai}
J.~W. Gichoya, I.~Banerjee, A.~R. Bhimireddy, J.~L. Burns, L.~A. Celi, L.-C. Chen, R.~Correa, N.~Dullerud, M.~Ghassemi, S.-C. Huang \emph{et~al.}, ``Ai recognition of patient race in medical imaging: a modelling study,'' \emph{The Lancet Digital Health}, vol.~4, no.~6, pp. e406--e414, 2022.

\bibitem{yang2023adversarial}
J.~Yang, A.~A. Soltan, D.~W. Eyre, Y.~Yang, and D.~A. Clifton, ``An adversarial training framework for mitigating algorithmic biases in clinical machine learning,'' \emph{NPJ digital medicine}, vol.~6, no.~1, p.~55, 2023.

\bibitem{xian2024optimal}
R.~Xian and H.~Zhao, ``Optimal group fair classifiers from linear post-processing,'' \emph{arXiv preprint arXiv:2405.04025}, 2024.

\bibitem{di2024post}
F.~Di~Gennaro, T.~Laugel, V.~Grari, X.~Renard, and M.~Detyniecki, ``Post-processing fairness with minimal changes,'' \emph{arXiv preprint arXiv:2408.15096}, 2024.

\bibitem{ktena2024generative}
I.~Ktena, O.~Wiles, I.~Albuquerque, S.-A. Rebuffi, R.~Tanno, A.~G. Roy, S.~Azizi, D.~Belgrave, P.~Kohli, T.~Cemgil \emph{et~al.}, ``Generative models improve fairness of medical classifiers under distribution shifts,'' \emph{Nature Medicine}, pp. 1--8, 2024.

\bibitem{meng2021knowledge}
H.~Meng, Z.~Lin, F.~Yang, Y.~Xu, and L.~Cui, ``Knowledge distillation in medical data mining: a survey,'' in \emph{5th International Conference on Crowd Science and Engineering}, 2021, pp. 175--182.

\bibitem{wu2024adaptive}
Z.~Wu and X.~Li, ``Adaptive knowledge distillation for high-quality unsupervised mri reconstruction with model-driven priors,'' \emph{IEEE Journal of Biomedical and Health Informatics}, 2024.

\bibitem{xing2021categorical}
X.~Xing, Y.~Hou, H.~Li, Y.~Yuan, H.~Li, and M.~Q.-H. Meng, ``Categorical relation-preserving contrastive knowledge distillation for medical image classification,'' in \emph{Medical Image Computing and Computer Assisted Intervention--MICCAI 2021: 24th International Conference, Strasbourg, France, September 27--October 1, 2021, Proceedings, Part V 24}.\hskip 1em plus 0.5em minus 0.4em\relax Springer, 2021, pp. 163--173.

\bibitem{joyce2011kullback}
J.~M. Joyce, ``Kullback-leibler divergence,'' in \emph{International encyclopedia of statistical science}.\hskip 1em plus 0.5em minus 0.4em\relax Springer, 2011, pp. 720--722.

\bibitem{vesti2003comparison}
E.~Vesti, C.~A. Johnson, and B.~C. Chauhan, ``Comparison of different methods for detecting glaucomatous visual field progression,'' \emph{Investigative ophthalmology \& visual science}, vol.~44, no.~9, pp. 3873--3879, 2003.

\bibitem{simonyan2014very}
K.~Simonyan, ``Very deep convolutional networks for large-scale image recognition,'' \emph{arXiv preprint arXiv:1409.1556}, 2014.

\bibitem{huang2017densely}
G.~Huang, Z.~Liu, L.~Van Der~Maaten, and K.~Q. Weinberger, ``Densely connected convolutional networks,'' in \emph{Proceedings of the IEEE conference on computer vision and pattern recognition}, 2017, pp. 4700--4708.

\bibitem{he2016deep}
K.~He, X.~Zhang, S.~Ren, and J.~Sun, ``Deep residual learning for image recognition,'' in \emph{Proceedings of the IEEE conference on computer vision and pattern recognition}, 2016, pp. 770--778.

\bibitem{dosovitskiy2020image}
A.~Dosovitskiy, ``An image is worth 16x16 words: Transformers for image recognition at scale,'' \emph{arXiv preprint arXiv:2010.11929}, 2020.

\bibitem{beutel2017data}
A.~Beutel, J.~Chen, Z.~Zhao, and E.~H. Chi, ``Data decisions and theoretical implications when adversarially learning fair representations,'' \emph{arXiv preprint arXiv:1707.00075}, 2017.

\bibitem{he2022masked}
K.~He, X.~Chen, S.~Xie, Y.~Li, P.~Doll{\'a}r, and R.~Girshick, ``Masked autoencoders are scalable vision learners,'' in \emph{Proceedings of the IEEE/CVF conference on computer vision and pattern recognition}, 2022, pp. 16\,000--16\,009.

\bibitem{loshchilov2017decoupled}
I.~Loshchilov, ``Decoupled weight decay regularization,'' \emph{International Conference on Learning Representations}, 2019.

\end{thebibliography}

\end{document}